\pdfoutput=1

\documentclass[11pt]{article}

\usepackage[final]{EMNLP2023}

\usepackage{times, blindtext}
\usepackage{latexsym}
\usepackage{booktabs}
\usepackage{pgfplots,subcaption}
\usepackage{comment}
\usepackage[T1]{fontenc}

\usepackage[utf8]{inputenc}

\usepackage{microtype}
\usepackage{graphicx}
\usepackage{inconsolata}

\usepackage{makecell}
\usepackage[none]{hyphenat}
\usepackage[T1]{fontenc}
\usepackage{url}
\usepackage{multirow}
\usepackage{float}
\usepackage{graphicx}
\usepackage{cleveref}
\crefformat{section}{\S#2#1#3} 
\crefformat{subsection}{\S#2#1#3}
\crefformat{subsubsection}{\S#2#1#3}
\usepackage{xcolor}
\newcommand\saimucjacomments[1]{\newline\textcolor{blue}{saimucja says: "#1"}}
\renewcommand\saimucjacomments[1]{} 

\title{Retrieve and Copy: Scaling ASR Personalization to Large Catalogs}

\author{Sai Muralidhar Jayanthi$^*$ \\
        AWS AI Labs \\
         \texttt{saimucja@amazon.com}
        \And
  Devang Kulshreshtha$^*$ \\
        AWS AI Labs \\
         \texttt{kulshrde@amazon.com}
  \And
  Saket Dingliwal$^*$  \\
        AWS AI Labs \\
         \texttt{skdin@amazon.com}
          \AND 
    Srikanth Ronanki \\
        AWS AI Labs \\
         \texttt{ronanks@amazon.com}
        \And
    Sravan Bodapati \\
        AWS AI Labs \\
         \texttt{sravanb@amazon.com}
          }

\begin{document}

\maketitle
\def\thefootnote{*}\footnotetext{These authors contributed equally to this work}\def\thefootnote{\arabic{footnote}}
\begin{abstract}

Personalization of automatic speech recognition (ASR) models is a widely studied topic because of its many practical applications. Most recently, attention-based contextual biasing techniques are used to improve the recognition of rare words and/or domain specific entities. However, due to performance constraints, the biasing is often limited to a few 
thousand entities, restricting real-world usability. To address this, we first propose a “Retrieve and Copy” mechanism to improve latency while retaining the accuracy even when scaled to a large catalog. We also propose a training strategy to overcome the degradation in recall at such scale due to an increased number of confusing entities. Overall, our approach achieves up to 6\% more Word Error Rate reduction (WERR) and 3.6\% absolute improvement in F1 when compared to a strong baseline. Our method also allows for large catalog sizes of up to 20K without significantly affecting WER and F1-scores, while achieving at least 
20\% inference speedup per acoustic frame.

\end{abstract}


\section{Introduction}
\label{sec:intro}

End-to-end ASR models based on Connectionist Temporal Classification (CTC) \cite{graves2006connectionist} and Transducers \cite{graves2012sequence} are widely popular. 
Although these models have shown outstanding improvements over hybrid models, they often struggle to recognize uncommon domain-specific words. 
This is further exacerbated for streaming ASR models due to limited audio context \cite{51265}.
To tackle this problem, attention-based Contextual Adapters (CA) have been proposed to boost a list of custom entity words (called `catalog') 
and have showcased to work well with catalogs up to hundreds of catalog items \cite{sathyendra2022contextual, Dingliwal2022}. 
However, many industrial applications have larger catalogs that can comprise of tens of thousands of words for ASR personalization. For example, a catalog of products sold by a business, a list of customer names or a search in video-on-demand platforms.
In this work, we identify two main challenges in scaling the existing methods to larger lists:
(1) Computing attention scores for each catalog item can significantly increase the latency of the system (and often redundant!), which is prohibitively critical for any streaming application,
(2) Large catalogs have more phonetically similar words which makes it hard for the CA models to disambiguate the correct entity for boosting.

To address these challenges, we propose novel inference and training strategies.
Through our inference method called "Retrieve and Copy", we first retrieve a smaller subset of relevant entities using Approximate Nearest Neighbor (ANN) search from the large catalog and then use only the retrieved entities for contextual biasing. 
Our best model leverages Fast AI Similarity Search (FAISS) \cite{johnson2019billion} that is designed for fast retrieval at scale.
Further, we introduce a  
fine-tuning strategy using hard negatives for the CA models. We use clustering to identify phonetically similar words from the training data and help the model learn to disambiguate between 
them.
Overall, the contributions of our work are:
\begin{itemize}
    \itemsep-0.26em
    \item We propose "Retrieve and Copy" inference strategy for ASR personalization with large catalogs that achieves atleast 
    20\% inference speedup per acoustic frame without affecting accuracy for large catalogs.
    \item We propose a fine-tuning strategy for Contextual Adapters to better disambiguate between similar sounding custom entities to improve accuracy.
    \item Using different datasets and catalog types, we show that our proposed methods can scale upto 20K catalog items, resulting in up-to a 6\% more WERR and 3.6\% improvement in absolute F1 compared to a strong baseline.
\end{itemize}

\section{Related Work}
\label{sec:related-work}
Attention-based contextual biasing modules have widely been used by ASR systems to personalize 
towards a catalog of a few hundred custom entities \cite{pundak2018deep, bruguier2019phoebe, sathyendra2022contextual, Dingliwal2022, nam}. 
However, \citet{nam_plus} showed that inference latency increases significantly even with a few thousand catalog items.
Similar to our approach, they propose to filter a small set of entities using maximum inner product. However, their method reduces dependency of phrase-length in the attention computation for associative memory based biasing modules \cite{nam}. 
In contrast, we use a single vector to represent an entity and hence do not have this dependency. Also, their experiments are limited to catalogs of size 3K, while we 
scale to 20K custom entities because of retrieval methods like FAISS.
Further, we introduce a fine-tuning strategy that specifically tackles the challenges of large catalog size on accuracy. \citet{alon2019contextual} 
previously used difficult examples for ASR contextualization
but their methods relied on generating fuzzy alternatives using phonetic similarity metric, while \citet{nearest-neighbor-phrase} used an ANN search with audio features. On the other hand, we use a simple clustering based strategy that allows us to easily use the elements belonging to the same cluster as phonetically similar entities.

\section{Background}\label{sec:methodology-baseline}
A CTC encoder takes in an audio, passes it through multiple Conformer blocks \cite{gulati2020conformer}, and generates a sequence of word piece posteriors. 
Contextual Adapters (CA) is a separate module that is added to the CTC encoder for boosting custom entities for personalization. Let $X^{1:T}$ denote $T$ output audio feature vectors from the CTC encoder. Let $W^{1:N}$ be a list of $N$ custom entity words. CA comprises of two main components: (i) Catalog Encoder: an LSTM \cite{10.1162/neco.1997.9.8.1735} that encodes word-piece sequences of custom entities into vectors (denoted by $C^{1:N}$) (ii) Biasing Adapter: an attention module \cite{NIPS2017_3f5ee243} that uses $X^t$ for each audio frame $t \in [1, T]$ as query and $C^{1:N}$ as keys to generate biasing vectors $B^t$. $B^t$ is then added back to $X^t$, thereby boosting any relevant custom entity. Let $\theta^Q, \theta^K, \theta^V$ represent the query, key and value matrices of the Biasing Adapter respectively. Then for each time frame $t \in [1, T]$, attention operation is equivalent to finding a score of each entity word $W^n, n \in [1,N]$ using the inner product $s_t^n = \langle \theta^Q X^t, \theta^K C^n \rangle$ and then $B^t = \sum_1^N s_t^n \theta^V C^n$. 

For training this module, 
each 
audio-text pair $(x, y)$
is augmented with a list of boosting words 
$W$=$\{x, W'\}$,
wherein the word $w$ is 
from the ground truth transcript $y$ that has the least term-frequency in the entire training data 
and $W'$ is a random subset
of other low term-frequency words present in the training data
but not in $y$.
In this way, 
CA learns to distinguish word $w$ from the rest of words $W'$ and boost its probability in the output sequence. We choose words with lower term-frequency as they are the hardest ones to be 
recognized by the un-adapted CTC encoder model.
\section{Methodology}
\captionsetup{width=\textwidth}
\begin{figure*}[!htp]
    \centering
    \includegraphics[width=0.72\textwidth]{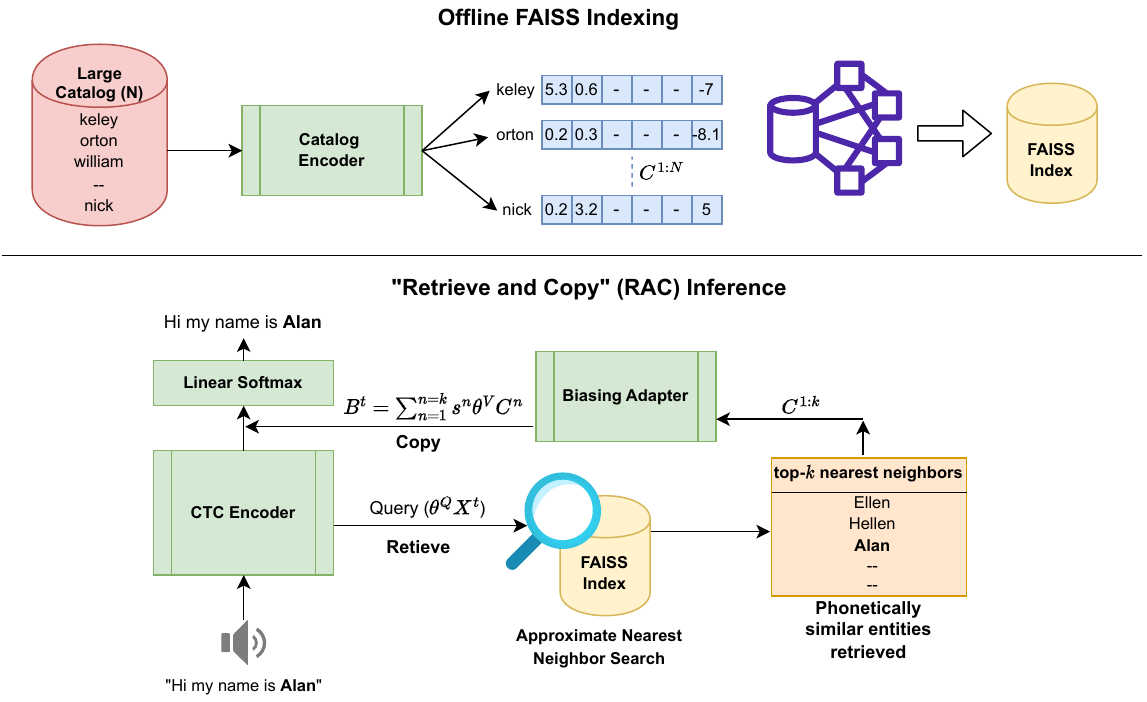}
    \vspace{-0.4cm}
    \caption{Details of our "Retrieve and Copy" inference strategy. (Top) Offline creation of FAISS index for a large catalog. (Bottom) Using ANN search to retrieve a subset of entities from the FAISS index for a given audio}
    \label{fig:faiss-inference}
\end{figure*}
\label{sec:methodology}
In many practical applications, the number of custom entities at inference time ($N$) can be substantially large and can contain up to 20K entities. As highlighted in Section \cref{sec:intro}, this creates challenges for both inference speed and performance. 
Following are our proposed inference- and training-side strategies designed to tackle these challenges respectively.
 
\subsection{Retrieve and Copy (RAC) Inference}
In order to reduce the inference latency, we need to find efficient ways to selectively reduce the catalog size to a smaller number at inference time. For this, we propose 
"Retrieve and Copy", where we first retrieve the most relevant entities for a given audio and then use 
them for 
CA.
Assuming 
either one or 
none of the custom entities will be spoken 
in a given 
audio,
the score $s_t^n$ of all but one would
be close to
0. 
Therefore, the biasing vector can be approximated using $B^t = \sum_1^k s_t^n \theta^V C^n$, where $k \ll N$ and $C^{1:k}$ are the vectors of the top-$k$ entities with the maximum inner product with the query vector ($\theta^Q X^t$) at any given time frame. This selection of top-$k$ entities reduces linear dependence of $N$ in the computation of attention in Contextual Adapters to $k$. We try different approaches for the retrieval of entities as summarized below.

\saimucjacomments{the above passage needs modifications in first few lines}

\noindent \textbf{Clustering}: We reduce the number of entities for biasing as follows: (i) cluster entities with similar vector representations,
(ii) choose most relevant cluster(s) for a given audio, (iii) use only the entities in the chosen cluster(s) for biasing. For the first step, we use $k$-means clustering on the vectors $\theta^K C^n$ using Euclidean distance to cluster $N$ entities into $M$ clusters ($M < N$) offline.
During inference, we score each cluster by computing the 
distance
between the query vector at each time frame and the centroid of the cluster.
Finally, we collect all the entities in each of the top-$l$ clusters and use them for biasing with Contextual Adapters. 



\noindent \textbf{Approximate Nearest Neighbors (ANN)}: In this approach, we leverage Theorem 1 in \citet{bachrach2014speeding} to transform the problem of finding top-$k$ entity vectors with maximum inner product 
to an ANN search problem. We transform our vectors from $d$-dimension to $d+1$ and find top-$k$ entities with 
the least Euclidean distance
with the query vector at each time frame.
Various methods have been proposed for solving ANN including FAISS and FAISS-IVF \cite{johnson2019billion},
and HNSWLIB \cite{malkov2018efficient}.\footnote{Tree based ANN methods such KDTree and BallTree \cite{scikit-learn}, and ANNOY \cite{annoy} haven't shown practical gains in inference latency and hence their results are excluded in Table~\ref{tab:main-results}} As shown in Figure \ref{fig:faiss-inference}, we create an index of our transformed custom entity vectors offline such that it can be efficiently queried for top-$k$ nearest neighbors during inference. At inference, for a given audio, we use the audio frame vector at each time step as the query, collect top-$k$ nearest neighbors, and then pass them to the Contextual Adapters for biasing. 






\subsection{Hard Negative Fine-tuning (HNFT)}\label{ssec:disc_training}
As the size of the catalog increases, we can find more phonetically similar entities within it, which makes it challenging for the Contextual Adapters to accurately disambiguate the correct entity.
Further, when our RAC inference strategy is applied, the set of top-$k$ retrieved entities, used for biasing are actually nearest neighbors in the Euclidean space. On the other hand, $w$ and $W'$ used during training are unrelated. This creates a mismatch between training and inference with large catalogs.  

In this work, we propose an additional fine-tuning stage for the Contextual Adapters to train them with similar-sounding words. 
We take all the low term-frequency words from CA training data (Section~\cref{sec:methodology-baseline}), pass them through the already trained CA Catalog Encoder and do $k$-means clustering into $s$ clusters. 
Then during fine-tuning, for an audio-text pair with the low term-frequency word $w$, instead of choosing a random subset of words $W'$, we choose words from the same cluster as $w$ as hard negatives. In this way, model learns to disambiguate between 
similar sounding words. 
Table \ref{tab:qualitative-clustering} showcases some training words belonging to the same cluster.

\label{ssec:expts-model-arch}


\captionsetup{width=.5\textwidth}
\begin{table}[!ht]
    \resizebox{\columnwidth}{!}{  
        \begin{tabular}{c|l}
            \toprule
            \textbf{ID} & \textbf{Training words in the same cluster} \\
            \toprule
            98 & bowl's, bolt's, bolz, bolles, bowell's, boby, boaby  \\
            \hline
            112 & froing, froning, refrying, refering, furloughing \\
            \hline
            234 & quake-hit, well-knit, top-knot, k-cup, pay-cuts, pay-cut  \\
            \hline
            999 & conjoining, congenial, conjugal, convivial, conjuncture \\
            \bottomrule
        \end{tabular}
    }
    \caption{Words from randomly picked clusters for HNFT} 
    \label{tab:qualitative-clustering}
\end{table}

\section{Experimental Setup}
\label{sec:expts}

\subsection{Evaluation Datasets}
\label{ssec:expts-test}

We extensively test our approach on five in-house conversational datasets and a public dataset. The details of the datasets are provided below and summarized in the Table \ref{tab:dataset-stats}.

\noindent \textbf{First Names \& Last Names}: Each utterance of this dataset contains a speaker telling their first name or last name respectively. In addition, they consist of a carrier phrase (CP) such as "my name is", "my first is", "yeah it is", etc. along with the name. We use a list of 20K common first and last names as catalog for these datasets respectively. 
We create random subsets from the large catalog consisting of all ground truth entities to carry out experiments related to varying catalog sizes.

\noindent \textbf{First Names w/o CP \& Last Names w/o CP}: These datasets are similar to First Names and Last names except they do not contain any carrier phrases. Again, we individually use a list of 20K names for each of these datasets. 

\noindent \textbf{Occupation}: Each utterance of this dataset contains a speaker telling about their occupation such as "cinematographer", "mammographer", etc. They may use a long or a short carrier phrase along with the occupation typical of a conversational setting. We use a list of 9K common occupations as catalog for this dataset.

\noindent \textbf{VoxPopuli} \cite{wang2021voxpopuli} is a public dataset of European Parliament event recordings from which we use the English test partition for our experiments.
This dataset contains long audios and the entity words compose a very small percentage of total words in the dataset.
For this dataset, we create an in-house catalog consisting of first, last, city and country names as well as 92 rarest words in training split of Voxpopuli as measured against training data's term frequencies. 

\captionsetup{width=\columnwidth}
\begin{table}[!ht]
    \setlength{\tabcolsep}{1.1pt} 
    \resizebox{\columnwidth}{!}{  
        \begin{tabular}{l|c|c|c|c|c}
        \toprule
        \makecell[c]{\multirow{2}{*}{\textbf{Dataset}}}  & \textbf{Num} & \textbf{Avg. Audio} & \textbf{Num} & \textbf{Catalog} & \textbf{Ground Truth} \\
         & \textbf{Audios} & \textbf{Length (s)} & \textbf{Words} & \textbf{Size} & \textbf{Entities Size} \\
         
        \hline
        First Names  & 250 & 4.9 & 818 & 20K & 250 \\
        First Names w/o CP  & 250 & 4.3 & 250 & 20K & 250 \\
        Last Names & 250 & 5.2 & 821 & 20K & 250 \\
        Last Names w/o CP  & 250 & 4.7 & 250 & 20K & 250 \\
        Occupation  & 2160 & 4.9 & 19814 & 9K & 144 \\
        Voxpopuli  & 1842 & 9.6 & 44830 & 20K & 156 \\
        \bottomrule
        \end{tabular}
    }
    \caption{Statistics of different evaluation datasets}
    \label{tab:dataset-stats}
\end{table}

\subsection{Evaluation Metrics}
\label{ssec:eval-metrics-expts-test}
We report Word Error Rate Reduction (WERR) (\%) on entire dataset
and F1 scores (\%) 
of the ground truth entities as evaluation metrics. 
Ground truth entities are those that are present in both large catalog as well as test set transcripts.
When computing inference latencies, we compute the wall clock time overhead of the contextual adapters module attached to the streaming ASR (Section~\cref{sec:methodology-baseline}) model in milliseconds (ms) per audio on a single CPU machine without multi-processing. 

\subsection{Models}
\label{ssec:model-architecture}

We evaluate three models -- streaming ASR model 
without contextual adapters (\textbf{Baseline}) and with contextual adapters (\textbf{CA}),
and a model further tuned with our proposed finetuning strategy in Section~\cref{ssec:disc_training} (\textbf{HNFT}).

Our models are trained with ESPnet \cite{watanabe2018espnet} using Conformer blocks \cite{gulati2020conformer} and joint CTC-Attention framework \cite{kim2017joint, watanabe2017hybrid} with Adam optimizer \cite{kingma2014adam}. The Baseline model is trained with 50K+ hours of speech-text parallel corpus in English.
For CA \& HNFT models, we follow 
\citet{Dingliwal2022}'s proposal and curate a subset of 1K hour from the parallel corpus leading to 230K catalog entities for adapting. 
For HNFT, we cluster these 230K catalog based on their embeddings from CA model into $s=1000$ clusters.
For inference, we finalize the RAC hyper-parameters ($k, M, l$) based on the performance of First Names dataset. 
Further, we train a 4-gram language model (LM) using the parallel corpus's texts for shallow fusion \cite{8462682} during ASR decoding. We refer the reader to Appendix~\ref{sec:model-training} for more details on model training and implementation.

\section{Results}\label{sec:results}

\captionsetup{width=.5\textwidth}
\begin{table}[!ht]
    \resizebox{\columnwidth}{!}{  
        \begin{tabular}{l|l|cc|cc}
        \hline
         & \makecell[c]{\multirow{2}{*}{\textbf{Methodology}}} & \multicolumn{2}{c|}{\textbf{First Names}} & \multicolumn{2}{c}{\textbf{Last Names}} \\
         & & \multirow{1}{*}{\textbf{F1}} & \textbf{Lat (ms)} & \multirow{1}{*}{\textbf{F1}} & \textbf{Lat (ms)} \\ 
        \hline
        1 & Baseline & 71.4 & N/A & 67.9 & N/A \\
        2  &\ \ + CA & 73.0 & 75.9 & 74.8 & 84.7 \\
        \hline
        3& \ \ \ \ + Clustering & 73.0 & 71.6 & 73.5 & 64.8 \\
        4& \ \ \ \ + HNSWLIB & 73.6 & 62.2 & 75.4 & 57.7 \\
        5 & \ \ \ \ + FAISS-IVF & 72.8 & \textbf{50.6} & 73.6 & \textbf{52.3} \\
        6 & \ \ \ \ + FAISS & \textbf{73.6} & 60.7 & \textbf{75.4} & 57.5 \\
        \bottomrule
        \end{tabular}
    }
    \caption{Performance comparison of Baseline, CA and different retrieval methods of our proposed RAC inference}
    
    \label{tab:main-results}
\end{table}

\captionsetup{width=\textwidth}
\begin{table*}[!ht]
    \small
    \centering
    \setlength{\tabcolsep}{2.0pt} 
    \resizebox{\textwidth}{!}{  
        \tabcolsep=0.11cm
        \begin{tabular}{l|l|cc|cc|cc|cc|cc|cc}
        \toprule
         & \makecell[c]{\multirow{2}{*}{\textbf{Model}}} &
          \multicolumn{2}{c|}{\textbf{First Names}} &
          \multicolumn{2}{c|}{\textbf{Last Names}} &
        \multicolumn{2}{c|}{\textbf{First Names w/o CP}} &
          \multicolumn{2}{c|}{\textbf{Last Names w/o CP}} &
          \multicolumn{2}{c|}{\textbf{Occupation}} &
          \multicolumn{2}{c}{\textbf{Voxpopuli}} \\
          & & \textbf{WERR (\%)} & \textbf{F1(\%)} & \textbf{WERR (\%)} & \textbf{F1(\%)} & \textbf{WERR(\%)} & \textbf{F1(\%)} & \textbf{WERR(\%)} & \textbf{F1(\%)} & \textbf{WERR(\%)} & \textbf{F1(\%)} & \textbf{WERR(\%)} & \textbf{F1(\%)}\\ 
          \toprule
          & \multicolumn{11}{l}{w/o Retrieve} \\
          \cmidrule{1-14}
          1 & Baseline & 0.0 & 71.4 & 0.0 & 67.9 & 0.0 & 68.6 & 0.0 & 54.7 & 0.0 & 54.8 & \textbf{0.0} & 65.5 \\  
          2 & \ \ + CA & 1.1 & 73.0 & 13.1 & 74.8 & 4.2 & 70.1 & 10.0 & 60.0 & \textbf{0.4} & 63.2 & -2.6 & 69.5 \\ 
          3 & \ \ \ \ + HNFT & 5.3 & 75.3 & 13.0 & 77.5 & 7.3 & 71.6 & 15.0 & 62.5 & 0.2 & \textbf{65.2} & -5.5 & \textbf{71.5} \\ 
          \toprule
          & \multicolumn{11}{l}{Retrieve (FAISS)} \\
          \cmidrule{1-14}
          4 & \ \ + CA & 1.6 & 73.6 & \textbf{14.0} & 75.4 & 5.8 & 70.8 & 10.0 & 60.0 & 0.1 & 63.5 & -2.7 & 70.3 \\ 
          5 & \ \ \ \ + HNFT & \textbf{5.9} & \textbf{75.9} & 11.6 & \textbf{77.8} & \textbf{8.1} & \textbf{71.9} & \textbf{16.0} & \textbf{63.6} & -1.2 & 65.1 & -5.6 & \textbf{71.5} \\ 
         \toprule
        \end{tabular}
    }
    \vspace{-0.2cm}
    \caption{WERR (\%) and F1 (\%) scores for models described in Section \ref{ssec:model-architecture} with and w/o retrival based inference. The WER of our Baseline model for VoxPopuli dataset is 10.5, in line with streaming models of similar sizes.}
    \label{tab:training-strategies}
\end{table*}
\captionsetup{width=\textwidth}
\newcommand{\rulesep}{\unskip\ \vrule\ }
\begin{figure*}[htb]
    \centering 
\begin{subfigure}{0.32\textwidth}
  \includegraphics[width=\linewidth]{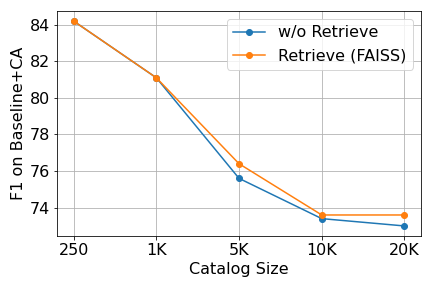}
  \label{fig:f1_configs}
\end{subfigure}
\rulesep
\begin{subfigure}{0.32\textwidth}
  \includegraphics[width=\linewidth]{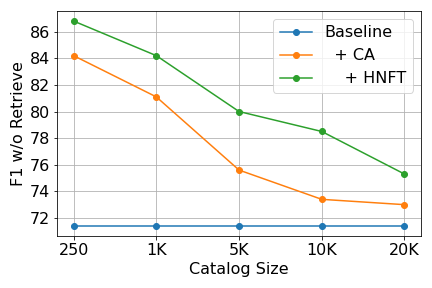}
  \label{fig:f1_models_fullcatalog}
\end{subfigure}
\rulesep
\begin{subfigure}{0.32\textwidth}
  \includegraphics[width=\linewidth]{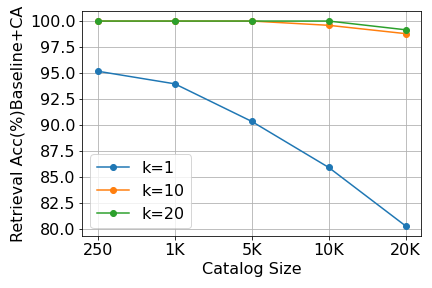}
  \label{fig:faiss_retrieval_accuracy}
\end{subfigure}
\begin{subfigure}{0.32\textwidth}
  \includegraphics[width=\linewidth]{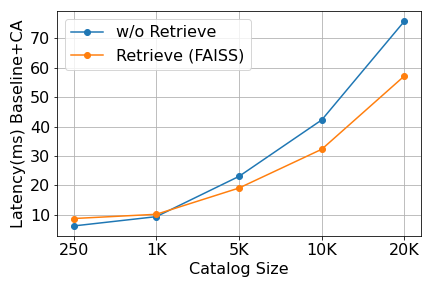}
  \caption{}
  \label{fig:latency_configs}
\end{subfigure} 
\rulesep
\begin{subfigure}{0.32\textwidth}
  \includegraphics[width=\linewidth]{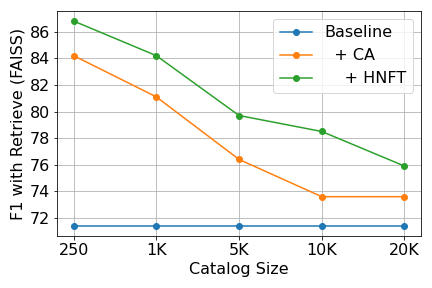}
  \caption{}
  \label{fig:f1_models_faiss}
\end{subfigure} 
\rulesep
\begin{subfigure}{0.32\textwidth}
  \includegraphics[width=\linewidth]{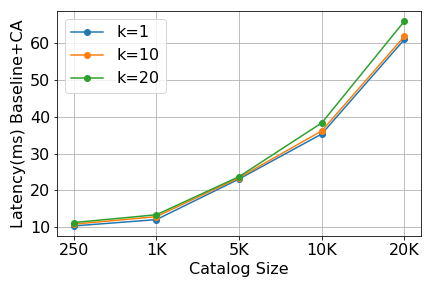}
  \caption{}
  \label{fig:faiss_latency}
\end{subfigure}
\caption{\textbf{(a)} Effect of RAC inference on F1 (top) and Latency (bottom) for varying catalog sizes. \textbf{(b)} Effect of HNFT on F1 without (top) and with RAC (bottom) for varying catalog sizes. \textbf{(c)} Effect of hyper-parameter $k$ in FAISS on retrieval accuracy (top) and latency (bottom) for varying catalog sizes.}
\label{fig:ablation_catalog_sizes}
\end{figure*}
\noindent \textbf{RAC Inference achieves the lowest inference latency with no performance regression}: In Table \ref{tab:main-results}, 
we compare the inference latencies and F1 scores in the retrieval of entity words on two of our datasets.
We compare the Baseline model, CA with and without our RAC inference strategy,
providing
a comparison of different retrieval methods. First, we observe that our proposed inference strategy yields significant improvements in latency over the standard inference with large catalog. Particularly, our best retrieval method FAISS reduces latency by 20-32\% (75.9ms to 60.7ms and 84.7ms to 57.5ms) compared to the inference without retrieval (Row 2 v/s 6).
Second, we see that this speedup doesn't come at the cost of performance, as our FAISS-based method achieves the best F1 score for both the datasets. This means the entities retrieved by FAISS based ANN search almost always contain the correct entity. We confirm this hypothesis later in our experiments. In fact, we observe some minor improvements in performance due to possible removal of unrelated entities in the retrieval step. Finally, among all the retrieval methods chosen for our experiment, FAISS ($k=10$) performs the best. On the other hand, Clustering ($M=2000, l=4$) performs worse than ANN based retrieval methods (Row 3) in both performance and latency. 
Due to it's performance,  FAISS will be our choice for retrieval in all the subsequent experiments. We further validate our claim of retaining the performance (or even improving) when we use our proposed efficient inference strategy on more datasets in the Table \ref{tab:training-strategies} (Row 2 v/s 4, Row 3 v/s 5).

\noindent \textbf{HNFT improves the WERR/F1 of Contextual Adapters}:
In Table \ref{tab:training-strategies}, we present an extensive comparison of our proposed training approach against baseline on different datasets. Comparing F1 scores (Row 2 v/s 3, Row 4 v/s 5),  we note that our fine-tuning strategy using hard negatives outperforms CA in biasing the correct custom entity for all the datasets. 
The WERR (\%) also improves or remains similar for all the datasets. For Voxpopuli,  while the F1 score improves significantly, there are WERR regressions with the use of CA with a large catalog. This is because the custom words are only $<0.2\%$ of total number of words in the dataset. While CA can recognize the right entity word (which are typically the most important words of the utterance), they sometimes unnecessarily substitute common words. This results in an increase in overall WER, which is in line with previous findings on the use of contextual biasing  \cite{nam_plus}.  In Table \ref{tab:qualitative_models}, we show qualitative examples of the output of our models on First Names dataset. The CA boosts either a phonetically similar word (\textit{Ruben}) or struggle to boost any entity word (\textit{Wy}) from the catalog. However, once we train our model with phonetically similar entities, it can disambiguate the subtle difference between these entities and can recognize the correct entity (\textit{Rueben} and \textit{Wally}).
\captionsetup{width=\columnwidth}
\begin{table}[]\small
    \centering
    \begin{tabular}{l|l}
    \hline
    \makecell[c]{\textbf{Model}} & \makecell[c]{\textbf{Transcription}} \\
    \hline
    Baseline & my name is \textcolor{red}{Ruben} \\
    \ \ + CA & my name is \textcolor{red}{Ruben} \\
     \ \ \ \ + HNFT  & my name is \color{green}{Rueben} \\
   \hline
    Baseline & my name is \textcolor{green}{Wally} \\
    \ \  + CA & yes it's \textcolor{red}{Wy} \\
    \ \ \ \  + HNFT  & yes it's \textcolor{green}{Wally} \\
    \hline
    \end{tabular}
    \vspace{-0.2cm}
    \caption{{Examples of generated transcripts for various models described in Section \ref{ssec:model-architecture} on First Names dataset.}}
    \label{tab:qualitative_models}
\end{table}


In Figure \ref{fig:ablation_catalog_sizes},  we study the effect of varying catalog size on different latency and accuracy metrics for our proposed methods. We use First Names dataset for the ablation and randomly subsample our total 20K catalog into subsets of smaller sizes (250, 1K, 5K and 10K). For each subset, all the ground truth entity words are  retained in the subset so that we can independently study the impact of the size of the catalog. 

\noindent \textbf{The latency improvements of RAC Inference over baseline increases with increase in catalog size}: In the Figure \ref{fig:latency_configs}, we compare the F1 and latency of our CA model with and without FAISS retrieval. We observe that there is a consistent decrease in F1 and an increase in latency as the catalog size increases (blue line). This validates our identified challenges of the problem of scaling contextual biasing methods to large catalogs. We also observe that our proposed FAISS based inference can help reduce the increase in latency with the increase in catalog size while maintaining similar F1 scores (orange line). Notably, the difference in the latency between our inference strategy and baseline increases with the increase in catalog size. This suggests that we can possibly go beyond catalogs of size 20K without a lot of increase in latency with our method. 

\noindent \textbf{HNFT improves accuracy consistently for all catalog sizes}: In Figure \ref{fig:f1_models_faiss}, we compare the 
F1 scores of our proposed fine-tuning with hard negatives strategy to Baseline and CA for different catalog sizes. We observe that the improvements of our method in the retrieval of the correct entity are consistently equal for different catalog sizes (orange vs green line). This holds for both with and without our RAC inference strategy (bottom and top sub-figures respectively). While our training method definitely improves the F1 scores, the benefits of using contextual biasing approach diminish over the Baseline for very large catalog sizes (blue line).


\noindent \textbf{top-$k$ ANN search can retrieve the correct entity with almost 100\% accuracy}: Finally, we try to understand the trade-off between latency and performance for different choices of our hyper-parameter $k$ in the Figure \ref{fig:faiss_latency}. In the top sub-figure, we plot retrieval accuracy, defined as the percentage of audios in which the correct entity was retrieved by top-$k$ ANN search using FAISS. Notably, for $k=1$, the accuracy drops significantly with large catalog size but we observe a very high retrieval accuracy (close to 100\%) for $k=10$ indicating that we do not lose any important entity as we select a subset of the large catalog to copy. In the bottom sub-figure, we observe that the latency is similar for different $k$. Hence, we choose $k=10$ as an optimum value for all our experiments. 




\saimucjacomments{also apple's technique drawback is that we need to tune n,k,append ratio which is a overhead for large models (see their fig.2)...our clustering approach has one parameter and keeping it close to the train catalog size is sufficient}

\saimucjacomments{layer normalization, proposed in NAM+ affects the performance as; mentioned we tested it out and find not having it is leading to best performance across datasets}
\section{Conclusions}
\label{sec:conclusions}
In this work, we identified the challenges with the use of contextual biasing methods for an industrial use case of biasing towards large catalogs. As the size of the catalog increases, we see a significant increase in latency and a corresponding drop in accuracy, making it practically infeasible to use existing approaches at such scale. To mitigate these challenges, we propose a "Retrieve and Copy" inference that leverages efficient ANN search methods like FAISS to selectively choose a small subset of relevant entities per audio, thereby improving inference latency by at least 20\%. Additionally, to improve the accuracy, we propose a fine-tuning strategy that uses phonetically similar words as hard negatives to train the model. It yields up to 6\% more WER reduction and up to 3.6\% absolute increase in F1 scores on one of our datasets.


\section*{Ethics Statement}
We hereby acknowledge that all of the co-authors of this work are aware of the provided ACL Code of Ethics and honor the code of conduct.
In this work, we focus on scaling personalization of ASR systems to large catalog lists using contextual biasing modules. For our experiments, we use a Baseline model trained using 50K+ hours of a large paired audio-text English data. Though large, we do not claim that this data is representative of all groups, accents and use cases. Our biasing mechanism can be effective in bridging the gap in the performance disparity for different groups by allowing for large custom lists. However, our methods and models are still susceptible to generating better outputs for certain groups of users. For example -  even a 20K list of first names might miss names from particular communities more than the others. Therefore, scaling beyond 20K entities might be necessary to make our method inclusive of a range of users and will be studied as part of future work.



\section*{Limitations}
Our RAC inference methodology improved the latency in scaling contextual biasing for large catalogs but we still see a consistent drop in F1 with increasing catalog size (\ref{fig:f1_models_faiss}). Incorporating hard negatives based fine-tuning helped, but more work is needed to scale our approach to even larger catalog size. 
Secondly, contextual biasing approaches can help in biasing the relevant entity but they can cause regressions on other common words in the dataset. In our experiments, we found that using CA on datasets with long audios like VoxPopuli can have WER regressions, specially with catalog of large size. This is another challenge in scaling these systems to some practical use cases that we plan to tackle in a future work.
Lastly, privacy and intellectual property concerns prevent us from releasing the training and evaluation datasets, limiting replication by other researchers.

\bibliography{strings,refs}

\begin{thebibliography}{27}
\expandafter\ifx\csname natexlab\endcsname\relax\def\natexlab#1{#1}\fi

\bibitem[{ann()}]{annoy}

\newblock {ANNOY} library.
\newblock \url{https://github.com/spotify/annoy}.
\newblock Accessed: 2023-07-15.

\bibitem[{Alon et~al.(2019)Alon, Pundak, and Sainath}]{alon2019contextual}
Uri Alon, Golan Pundak, and Tara~N Sainath. 2019.
\newblock Contextual speech recognition with difficult negative training examples.
\newblock In \emph{ICASSP 2019-2019 IEEE International Conference on Acoustics, Speech and Signal Processing (ICASSP)}, pages 6440--6444. IEEE.

\bibitem[{Bachrach et~al.(2014)Bachrach, Finkelstein, Gilad-Bachrach, Katzir, Koenigstein, Nice, and Paquet}]{bachrach2014speeding}
Yoram Bachrach, Yehuda Finkelstein, Ran Gilad-Bachrach, Liran Katzir, Noam Koenigstein, Nir Nice, and Ulrich Paquet. 2014.
\newblock Speeding up the xbox recommender system using a euclidean transformation for inner-product spaces.
\newblock In \emph{Proceedings of the 8th ACM Conference on Recommender systems}, pages 257--264.

\bibitem[{Bleeker et~al.(2023)Bleeker, Swietojanski, Braun, and Zhuang}]{nearest-neighbor-phrase}
Maurits Bleeker, Pawel Swietojanski, Stefan Braun, and Xiaodan Zhuang. 2023.
\newblock \href {https://arxiv.org/abs/2304.08862} {Approximate nearest neighbour phrase mining for contextual speech recognition}.
\newblock In \emph{Interspeech}.

\bibitem[{Bruguier et~al.(2019)Bruguier, Prabhavalkar, Pundak, and Sainath}]{bruguier2019phoebe}
Antoine Bruguier, Rohit Prabhavalkar, Golan Pundak, and Tara~N Sainath. 2019.
\newblock Phoebe: Pronunciation-aware contextualization for end-to-end speech recognition.
\newblock In \emph{ICASSP 2019-2019 IEEE International Conference on Acoustics, Speech and Signal Processing (ICASSP)}, pages 6171--6175. IEEE.

\bibitem[{Chiu et~al.(2021)Chiu, Cao, Siohan, Pang, Doutre, and Han}]{51265}
Chung-Cheng Chiu, Liangliang Cao, Olivier Siohan, Ruoming Pang, Thibault Doutre, and Wei Han. 2021.
\newblock Bridging the gap between streaming and non-streaming automatic speechrecognition systems through distillation of an ensemble of models.
\newblock In \emph{Interspeech'2021}.

\bibitem[{Dingliwal et~al.(2023)Dingliwal, Sunkara, Ronanki, Farris, Kirchhoff, and Bodapati}]{Dingliwal2022}
Saket Dingliwal, Monica Sunkara, Srikanth Ronanki, Jeff Farris, Katrin Kirchhoff, and Sravan Bodapati. 2023.
\newblock \href {https://doi.org/10.1109/SLT54892.2023.10022705} {Personalization of ctc speech recognition models}.
\newblock In \emph{2022 IEEE Spoken Language Technology Workshop (SLT)}, pages 302--309.

\bibitem[{Graves(2012)}]{graves2012sequence}
Alex Graves. 2012.
\newblock \href {http://arxiv.org/abs/1211.3711} {Sequence transduction with recurrent neural networks}.

\bibitem[{Graves et~al.(2006)Graves, Fern{\'a}ndez, Gomez, and Schmidhuber}]{graves2006connectionist}
Alex Graves, Santiago Fern{\'a}ndez, Faustino Gomez, and J{\"u}rgen Schmidhuber. 2006.
\newblock Connectionist temporal classification: labelling unsegmented sequence data with recurrent neural networks.
\newblock In \emph{Proceedings of the 23rd international conference on Machine learning}, pages 369--376.

\bibitem[{Gulati et~al.(2020)Gulati, Qin, Chiu, Parmar, Zhang, Yu, Han, Wang, Zhang, Wu et~al.}]{gulati2020conformer}
Anmol Gulati, James Qin, Chung-Cheng Chiu, Niki Parmar, Yu~Zhang, Jiahui Yu, Wei Han, Shibo Wang, Zhengdong Zhang, Yonghui Wu, et~al. 2020.
\newblock Conformer: Convolution-augmented transformer for speech recognition.
\newblock \emph{arXiv preprint arXiv:2005.08100}.

\bibitem[{Hochreiter and Schmidhuber(1997)}]{10.1162/neco.1997.9.8.1735}
Sepp Hochreiter and J\"{u}rgen Schmidhuber. 1997.
\newblock \href {https://doi.org/10.1162/neco.1997.9.8.1735} {Long short-term memory}.
\newblock \emph{Neural Comput.}, 9(8):1735–1780.

\bibitem[{Johnson et~al.(2019)Johnson, Douze, and J{\'e}gou}]{johnson2019billion}
Jeff Johnson, Matthijs Douze, and Herv{\'e} J{\'e}gou. 2019.
\newblock Billion-scale similarity search with {GPUs}.
\newblock \emph{IEEE Transactions on Big Data}, 7(3):535--547.

\bibitem[{Kannan et~al.(2018)Kannan, Wu, Nguyen, Sainath, Chen, and Prabhavalkar}]{8462682}
Anjuli Kannan, Yonghui Wu, Patrick Nguyen, Tara~N. Sainath, ZhiJeng Chen, and Rohit Prabhavalkar. 2018.
\newblock \href {https://doi.org/10.1109/ICASSP.2018.8462682} {An analysis of incorporating an external language model into a sequence-to-sequence model}.
\newblock In \emph{2018 IEEE International Conference on Acoustics, Speech and Signal Processing (ICASSP)}, pages 1--5828.

\bibitem[{Kim et~al.(2017)Kim, Hori, and Watanabe}]{kim2017joint}
Suyoun Kim, Takaaki Hori, and Shinji Watanabe. 2017.
\newblock {Joint CTC-attention based end-to-end speech recognition using multi-task learning}.
\newblock In \emph{Proc. ICASSP}, pages 4835--4839.

\bibitem[{Kingma and Ba(2014)}]{kingma2014adam}
Diederik~P Kingma and Jimmy Ba. 2014.
\newblock Adam: A method for stochastic optimization.
\newblock \emph{arXiv preprint arXiv:1412.6980}.

\bibitem[{Kudo and Richardson(2018)}]{kudo-richardson-2018-sentencepiece}
Taku Kudo and John Richardson. 2018.
\newblock \href {https://doi.org/10.18653/v1/D18-2012} {{S}entence{P}iece: A simple and language independent subword tokenizer and detokenizer for neural text processing}.
\newblock In \emph{Proceedings of the 2018 Conference on Empirical Methods in Natural Language Processing: System Demonstrations}, pages 66--71, Brussels, Belgium. Association for Computational Linguistics.

\bibitem[{Lee and Watanabe(2021)}]{lee2021intermediate}
Jaesong Lee and Shinji Watanabe. 2021.
\newblock {Intermediate loss regularization for CTC-based speech recognition}.
\newblock In \emph{Proc. ICASSP}, pages 6224--6228.

\bibitem[{Malkov and Yashunin(2018)}]{malkov2018efficient}
Yu~A Malkov and Dmitry~A Yashunin. 2018.
\newblock Efficient and robust approximate nearest neighbor search using hierarchical navigable small world graphs.
\newblock \emph{IEEE transactions on pattern analysis and machine intelligence}, 42(4):824--836.

\bibitem[{Munkhdalai et~al.(2022)Munkhdalai, Sim, Chandorkar, Gao, Chua, Strohman, and Beaufays}]{nam}
Tsendsuren Munkhdalai, Khe~Chai Sim, Angad Chandorkar, Fan Gao, Mason Chua, Trevor Strohman, and Françoise Beaufays. 2022.
\newblock \href {https://doi.org/10.1109/ICASSP43922.2022.9747726} {Fast contextual adaptation with neural associative memory for on-device personalized speech recognition}.
\newblock In \emph{ICASSP 2022 - 2022 IEEE International Conference on Acoustics, Speech and Signal Processing (ICASSP)}, pages 6632--6636.

\bibitem[{Munkhdalai et~al.(2023)Munkhdalai, Wu, Pundak, Sim, Li, Rondon, and Sainath}]{nam_plus}
Tsendsuren Munkhdalai, Zelin Wu, Golan Pundak, Khe~Chai Sim, Jiayang Li, Pat Rondon, and Tara~N. Sainath. 2023.
\newblock \href {https://doi.org/10.1109/SLT54892.2023.10023323} {Nam+: Towards scalable end-to-end contextual biasing for adaptive asr}.
\newblock In \emph{2022 IEEE Spoken Language Technology Workshop (SLT)}, pages 190--196.

\bibitem[{Pedregosa et~al.(2011)Pedregosa, Varoquaux, Gramfort, Michel, Thirion, Grisel, Blondel, Prettenhofer, Weiss, Dubourg, Vanderplas, Passos, Cournapeau, Brucher, Perrot, and Duchesnay}]{scikit-learn}
F.~Pedregosa, G.~Varoquaux, A.~Gramfort, V.~Michel, B.~Thirion, O.~Grisel, M.~Blondel, P.~Prettenhofer, R.~Weiss, V.~Dubourg, J.~Vanderplas, A.~Passos, D.~Cournapeau, M.~Brucher, M.~Perrot, and E.~Duchesnay. 2011.
\newblock Scikit-learn: Machine learning in {P}ython.
\newblock \emph{Journal of Machine Learning Research}, 12:2825--2830.

\bibitem[{Pundak et~al.(2018)Pundak, Sainath, Prabhavalkar, Kannan, and Zhao}]{pundak2018deep}
Golan Pundak, Tara~N Sainath, Rohit Prabhavalkar, Anjuli Kannan, and Ding Zhao. 2018.
\newblock Deep context: end-to-end contextual speech recognition.
\newblock In \emph{2018 IEEE spoken language technology workshop (SLT)}, pages 418--425. IEEE.

\bibitem[{Sathyendra et~al.(2022)Sathyendra, Muniyappa, Chang, Liu, Su, Strimel, Mouchtaris, and Kunzmann}]{sathyendra2022contextual}
Kanthashree~Mysore Sathyendra, Thejaswi Muniyappa, Feng-Ju Chang, Jing Liu, Jinru Su, Grant~P Strimel, Athanasios Mouchtaris, and Siegfried Kunzmann. 2022.
\newblock Contextual adapters for personalized speech recognition in neural transducers.
\newblock In \emph{ICASSP 2022-2022 IEEE International Conference on Acoustics, Speech and Signal Processing (ICASSP)}, pages 8537--8541. IEEE.

\bibitem[{Vaswani et~al.(2017)Vaswani, Shazeer, Parmar, Uszkoreit, Jones, Gomez, Kaiser, and Polosukhin}]{NIPS2017_3f5ee243}
Ashish Vaswani, Noam Shazeer, Niki Parmar, Jakob Uszkoreit, Llion Jones, Aidan~N Gomez, \L~ukasz Kaiser, and Illia Polosukhin. 2017.
\newblock \href {https://proceedings.neurips.cc/paper_files/paper/2017/file/3f5ee243547dee91fbd053c1c4a845aa-Paper.pdf} {Attention is all you need}.
\newblock In \emph{Advances in Neural Information Processing Systems}, volume~30. Curran Associates, Inc.

\bibitem[{Wang et~al.(2021)Wang, Riviere, Lee, Wu, Talnikar, Haziza, Williamson, Pino, and Dupoux}]{wang2021voxpopuli}
Changhan Wang, Morgane Riviere, Ann Lee, Anne Wu, Chaitanya Talnikar, Daniel Haziza, Mary Williamson, Juan Pino, and Emmanuel Dupoux. 2021.
\newblock {Voxpopuli: A large-scale multilingual speech corpus for representation learning, semi-supervised learning and interpretation}.
\newblock \emph{arXiv preprint arXiv:2101.00390}.

\bibitem[{Watanabe et~al.(2018)Watanabe, Hori, Karita, Hayashi, Nishitoba, Unno, {Enrique Yalta Soplin}, Heymann, Wiesner, Chen, Renduchintala, and Ochiai}]{watanabe2018espnet}
Shinji Watanabe, Takaaki Hori, Shigeki Karita, Tomoki Hayashi, Jiro Nishitoba, Yuya Unno, Nelson {Enrique Yalta Soplin}, Jahn Heymann, Matthew Wiesner, Nanxin Chen, Adithya Renduchintala, and Tsubasa Ochiai. 2018.
\newblock \href {https://doi.org/10.21437/Interspeech.2018-1456} {{ESPnet}: End-to-end speech processing toolkit}.
\newblock In \emph{Proceedings of Interspeech}, pages 2207--2211.

\bibitem[{Watanabe et~al.(2017)Watanabe, Hori, Kim, Hershey, and Hayashi}]{watanabe2017hybrid}
Shinji Watanabe, Takaaki Hori, Suyoun Kim, John~R Hershey, and Tomoki Hayashi. 2017.
\newblock {Hybrid CTC/attention architecture for end-to-end speech recognition}.
\newblock \emph{IEEE Journal of Selected Topics in Signal Processing}, 11(8):1240--1253.

\end{thebibliography}
\bibliographystyle{acl_natbib}

\appendix
\section{Model Training}
\label{sec:model-training}

Our models are trained with joint CTC-Attention framework \cite{kim2017joint, watanabe2017hybrid} and intermediate CTC regularization \cite{lee2021intermediate} with 20 layers of Conformer blocks \cite{gulati2020conformer} consisting of 8 attention heads and 512 hidden dimension. During inference, we discard the Attention head and use CTC decoder for transcript generation. We train the Baseline model with 50K+ hours of speech-text parallel corpus in English consisting of a mix of accents, speakers, sampling rates and background noise.  

All our models are trained with Adam optimizer \cite{kingma2014adam}. We train the Baseline model for 30 epochs and continue training the CA model from its last checkpoint for 50 epochs by freezing all but adapter parameters. We adopt curriculum training for the CA model by linearly increasing the biasing catalog size during training from 30 to a maximum of 200 in steps of 4 per epoch and using random negatives drawn from the pool of 230K catalog. We hypothesize that gradually expanding the catalog size can make the model more robust to large catalog settings. HNFT model is finetuned on top of the CA model for 10 epochs.\footnote{Our experiments indicate that training for more than 10 epochs has no significant impact}

We train a SentencePiece \cite{kudo-richardson-2018-sentencepiece} tokenizer with token size of 2048 for encoding transcripts. Further, we train a 4-gram language model (LM) using the parallel corpus's texts for shallow fusion \cite{8462682}. We keep the tokenizer and the LM same across all the models. During inference, we use a beam size of 50 and LM weight of 0.6 in all our experiments. Our work is implemented in the open-source toolkit ESPnet \cite{watanabe2018espnet}.

\end{document}